\documentclass{article}

\usepackage{iclr2027_conference,times}

\usepackage[utf8]{inputenc}
\usepackage[T1]{fontenc}
\usepackage{graphicx}
\usepackage{booktabs}
\usepackage{amsmath}
\usepackage{amssymb}
\usepackage{fvextra}
\usepackage{tabularx}
\usepackage{longtable}
\usepackage{array}
\usepackage{multicol}
\usepackage{multirow}
\usepackage{xspace}
\usepackage{xcolor}
\usepackage{microtype}
\usepackage{url}
\usepackage{hyperref}
\usepackage{tabularx}
\usepackage{wrapfig}
\hypersetup{hidelinks}
\usepackage{fontawesome5}
\usepackage[accsupp]{axessibility}

\newcommand{\name}{VLAA-GUI\xspace}
\newcommand{\ie}{\emph{i.e.}\xspace}
\newcommand{\eg}{\emph{e.g.}\xspace}

\newcommand{\vs}{\emph{vs.}\xspace}

\title{Knowing When to \texttt{STOP}, \texttt{RECOVER}, and \texttt{SEARCH}: A Modular Framework for GUI Automation}

\newcommand{\authorsep}{\hspace{1.6em}}
\author{%
  \parbox{\textwidth}{\centering
      Qijun Han\textsuperscript{1,*}\authorsep Haoqin Tu\textsuperscript{1,*}\authorsep
      Zijun Wang\textsuperscript{1}\authorsep Haoyue Dai\textsuperscript{2}\authorsep
      Yiyang Zhou\textsuperscript{3}\\[0.15em]
      Nancy Lau\textsuperscript{1}\authorsep Alvaro A. Cardenas\textsuperscript{1}\authorsep
      Yuhui Xu\textsuperscript{4}\authorsep Ran Xu\textsuperscript{4}\authorsep
      Caiming Xiong\textsuperscript{4}\\[0.15em]
      Zeyu Zheng\textsuperscript{5}\authorsep Huaxiu Yao\textsuperscript{3}\authorsep
      Yuyin Zhou\textsuperscript{1}\authorsep Cihang Xie\textsuperscript{1}\\[0.5em]
    {\small\mdseries \textsuperscript{1}UC Santa Cruz\quad \textsuperscript{2}CMU\quad
    \textsuperscript{3}UNC-Chapel Hill\quad
    \textsuperscript{4}Salesforce\quad \textsuperscript{5}UC Berkeley}\\[0.3em]
    {\small\mdseries \textsuperscript{*}Equal contribution}
  }
}

\iclrfinalcopy
\makeatletter
\renewcommand{\@maketitle}{%
  \vspace*{-6em}
  {\centering
    \includegraphics[width=0.1\textwidth]{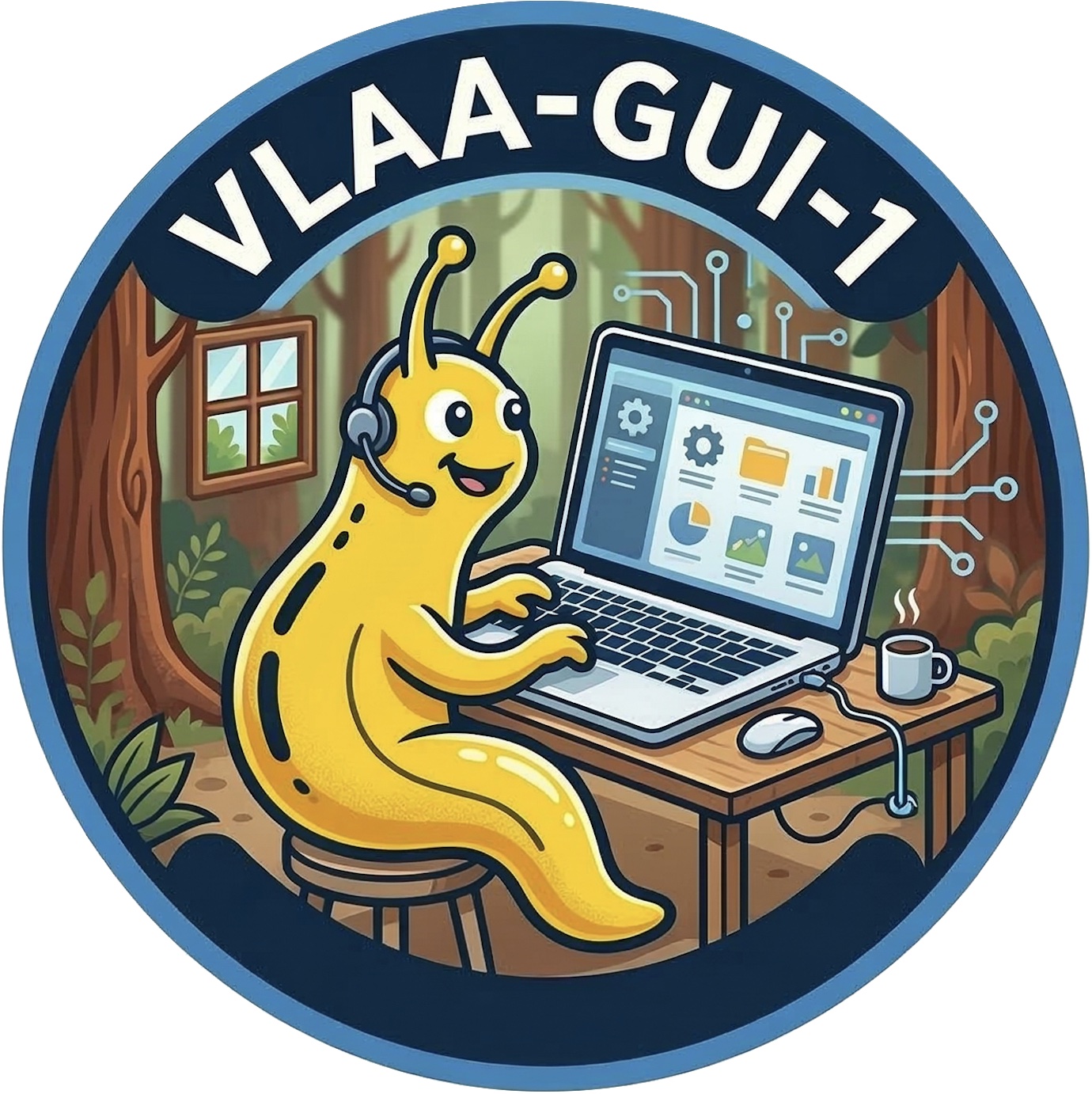}\par
    {\LARGE\scshape\@title\par}}
  {\centering\bfseries\@author\par}
}
\makeatother
\begin{document}

\maketitle
\vspace{0.2em}
{\centering\small
  \href{https://ucsc-vlaa.github.io/VLAA-GUI}{\faGlobe\ Website}%
  \;\;$\bullet$\;\;%
  \href{https://github.com/UCSC-VLAA/VLAA-GUI}{\faGithub\ Code}\par}
  \vspace{0.5em}

\begin{abstract}
Autonomous GUI agents face two fundamental challenges: \emph{early stopping}, where agents prematurely declare success without verifiable evidence, and \emph{repetitive loops}, where agents cycle through the same failing actions without recovery. 
Existing agent planners track progress through generated subtasks, yet in GUI environments, completing planned steps does not guarantee that the user's requirements are met. Moreover, existing anti-loop mechanisms typically detect only local repetition and lack mechanisms to escalate recovery when the agent remains stuck.
To this end, we present \textbf{\name}, a modular GUI agentic framework organized around three decisions: when to \texttt{Stop}, \texttt{Recover}, and \texttt{Search}. 
To prevent premature termination, a mandatory \emph{Completeness-Aware Planner} tracks UI-observable success criteria throughout execution and independently verifies completion claims before termination. 
To recover from stalled agent action loops, a mandatory \emph{Loop Breaker} detects repeated ineffective agent actions and progressively forces the agent to change its action strategy.
When the current observation is insufficient to determine a viable workflow, an on-demand \emph{Search Agent} supplies external procedural knowledge to the agent.
To validate \name, we evaluate it on two benchmarks containing Linux and Windows tasks with five LLM backbones, achieving top performance on both (\textbf{77.5\%} on OSWorld and \textbf{61.0\%} on WindowsAgentArena). Notably, three of the five backbones surpass human performance (72.4\%) on OSWorld in a single pass. 
Ablation studies further show that completion checking, loop recovery, and search provide complementary gains that vary with backbone capability and interaction budget, while the Loop Breaker nearly halves wasted steps for loop-prone backbones (\eg, Gemini 3 Flash). Moreover, transferring three proposed mechanisms to an existing GUI agent (CoAct-1) also improves it by over 5\% to 70.3\% on a controlled OSWorld subset, showing the strong portability of these components.
\end{abstract}

\section{Introduction}
\label{sec:intro}
\begin{figure}[!t]
    \centering
    \includegraphics[width=1\linewidth]{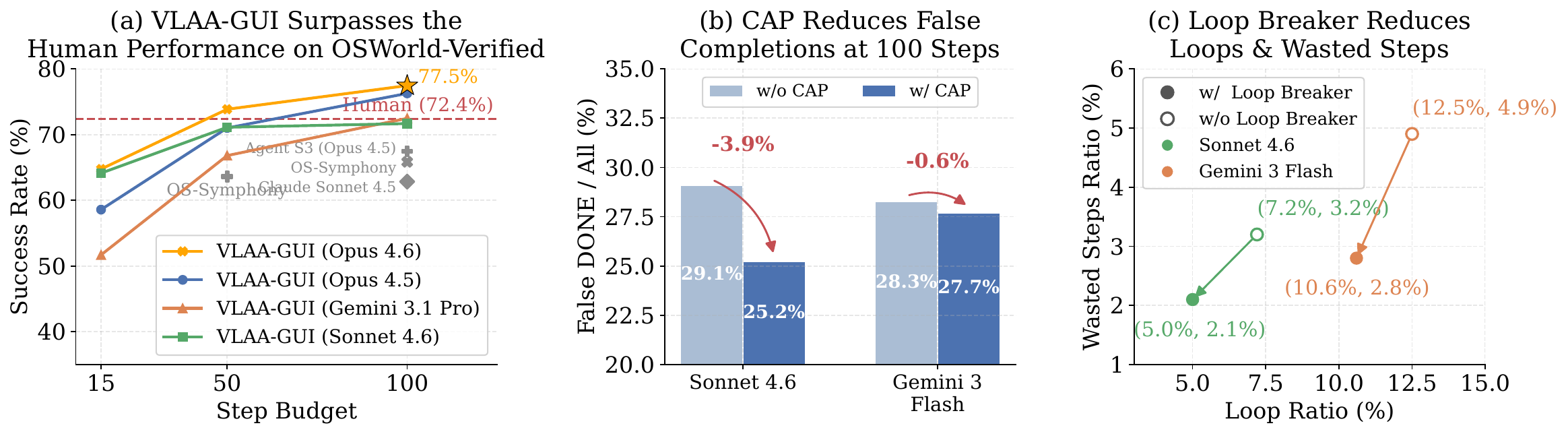}
     \caption{\textbf{Performance of \name.} (a) Our \name (w/ Opus 4.6) achieves the best results (77.5\%) on OSWorld-Verified~\citep{xie2024osworld} and surpasses human performance with one pass. (b) Completeness-Aware Planner (CAP) reduces false completion rates at a 100-step budget. (c) The Loop Breaker reduces repetitive loops and wasted action steps at the 100-step budget.}
     \vspace{-1.5em}
     \label{fig:teaser}
\end{figure}

The rapid advancement of multimodal large language models (MLLMs) has catalyzed a new generation of autonomous Graphical User Interface (GUI) 
agents~\citep{qin2025uitars,agashe2024agents,wang2025opencua,wu2024osatlas,hong2024cogagent} capable of performing desktop tasks by observing screenshots and executing mouse and keyboard actions. Systems such as OSWorld~\citep{xie2024osworld}, WindowsAgentArena~\citep{bonatti2024waa}, and related benchmarks~\citep{zhang2024llmguisurvey, yang2025macosworld} have established standardized evaluation environments spanning Linux, Windows and macOS, revealing both promises and persistent limitations of current approaches.

Despite steady progress, two fundamental problems remain largely unsolved in this domain. First, GUI agents do not reliably know \emph{when a task is finished}: they routinely declare success prematurely, \eg, after toggling a setting without verifying the state changed---because completion is left to the model's implicit judgment rather than verified against observable UI evidence~\citep{andrade2025selfgrounded,cemri2025multiagentfail}. Existing planning modules primarily focus on how to complete intermediate subgoals, but provide limited mechanisms for verifying whether the requested final state has actually been achieved~\citep{agashe2024agents,agashe2025agents2}.
Second, agents fall into \emph{repetitive loops}: cycling through the same failing action without recovery. Existing anti-looping heuristics can detect repeated actions, but often lack a systematic way to change strategy when repeated attempts continue to fail~\citep{yao2023react,shinn2023reflexion,madaan2023selfrefine,li2023structuredreflection,kim2023computertasks}.

We present \textbf{\name}, a modular GUI agent framework that addresses both challenges through three mechanisms, \emph{a Completeness-Aware Planner, a Loop Breaker, and a Search Agent} that enable the system to determine when to \texttt{STOP}, \texttt{RECOVER} (from repetitive loops), and \texttt{SEARCH} online for enhanced performance.
To mitigate early stopping, we further introduce the \emph{Completeness-Aware Planner} (CAP), which maintains a set of UI-observable completion criteria throughout execution and uses the status of unmet criteria at each step to guide subsequent decisions. Unlike conventional planners that prescribe a sequence of intermediate steps, CAP lets the agent choose actions dynamically based on the current UI state and what remains to be completed. This design helps prevent the agent from stopping simply because it has finished its planned actions while the requested final state has not yet been achieved.
To address repetitive loops, we introduce a mandatory \emph{Loop Breaker} that runs after every action and escalates its intervention as evidence of a loop accumulates---from switching the interaction modality (\eg clicking screen elements or writing Bash commands), to forcing an outright strategy change, to acting on an external loop signal---so that the agent is redirected before repeated failures consume the step budget. 

Beyond the two core challenges, GUI agents frequently stall on unfamiliar application workflows, leading to poor out-of-distribution performance. 
Prior work addresses this via visual browser-based search~\citep{yang2026ossymphony}, which requires additional agent steps and a visual grounding pipeline. We instead deploy an on-demand Search Agent that directly queries a capable LLM with search grounding (\eg, Gemini~3~Pro~\citep{gemini3pro2025}), returning plain textual knowledge injected into the Manager's context, avoiding the need to navigate a browser to retrieve procedural text. 
We further integrate a Coding Agent (Coder) for accurate command line executions and a Grounding Agent (Grounder) for visual grounding following existing works~\citep{gonzalez2025agents3,yang2026ossymphony,wang2024codeact}.
 
We evaluate \name on OSWorld-Verified~\citep{xie2024osworld} and WindowsAgentArena (WAA)~\citep{bonatti2024waa} across five LLM backbones to systematically validate the framework across distinct model families and capability tiers.
As shown in Figure~\ref{fig:teaser}, \name reaches 77.5\% with Opus~4.6, and Opus~4.5 and Gemini~3.1~Pro also exceed the human reference level of 72.4\%, making \name the first GUI framework to do so. Furthermore, \name generalizes well to WAA, reaching 61.0\% and outperforming the strongest baselines by over 4\%.

Ablation studies further validate the contribution of each component. The full \name system with Sonnet\~4.6 achieves 71.67\% on OSWorld, while removing the Completeness-Aware Planner, Searcher, or Loop Breaker reduces performance to 68.81\%, 70.04\%, and 71.63\%, respectively. To further examine whether these proposed modules are transferable to other GUI frameworks, we integrate them into CoAct-1~\citep{song2025coact}. On a fixed OSWorld subset, adding the Completeness-Aware Planner, Search Agent, and Loop Breaker individually improves CoAct-1 by 3.78\%, 0.83\%, and 0.93\%, respectively, while combining all three improves performance by over 5\% (\ie, 65.13\% to 70.34\%), highlighting their strong potential to transfer to other GUI frameworks.

\section{Related Work}
\label{sec:related}

\paragraph{GUI Agents: Models and Frameworks.}
End-to-end models trained for GUI interaction---UI-TARS~\citep{qin2025uitars}, AGUVIS~\citep{xu2025aguvis}, ShowUI~\citep{lin2024showui}, CogAgent~\citep{hong2024cogagent}, OS-Atlas~\citep{wu2024osatlas}, among others~\citep{yang2024ariaui,wang2025opencua,fu2025mano}---achieve strong grounding without HTML or accessibility trees. 
Frontier providers have followed with commercial APIs: Claude Computer Use~\citep{anthropic2024computeruse}, OpenAI CUA~\citep{openai2025cua}, and Seed-1.8~\citep{seed18}, the latter serving as our dedicated grounding model. Complementary modular frameworks combine MLLMs with planning, memory, and tools. The Agent~S family~\citep{agashe2024agents,agashe2025agents2} couples hierarchical planning with experience memory; Agent~S3~\citep{gonzalez2025agents3} adds best-of-N trajectory selection. OS-Symphony~\citep{yang2026ossymphony} and GTA1~\citep{yang2025gta1} combine memory systems and test-time search, while CoAct-1~\citep{song2025coact} and EvoCUA~\citep{xue2026evocua} emphasize coding-as-actions and synthetic experience. UFO~\citep{zhang2024ufo} and AutoGLM~\citep{liu2024autoglm} target Windows and mobile respectively. 

\paragraph{Self-Verification, Termination, and Error Recovery.}
Reliable termination remains difficult because progress must be inferred from partial, noisy observations. ReAct~\citep{yao2023react} and Tree of Thoughts~\citep{yao2023tree} enable mid-trajectory correction, while Reflexion~\citep{shinn2023reflexion} and Self-Refine~\citep{madaan2023selfrefine} improve policies through explicit self-feedback. For computer-control agents, structured reflection~\citep{li2023structuredreflection,kim2023computertasks} has been explored to reduce repeated failures. Complementarily, verifier-based training and step-wise checking~\citep{cobbe2021gsm8k,lightman2024lets} improve reliability in reasoning tasks; Pan~et~al.~\citep{pan2024autonomous} extend autonomous evaluation to digital agents. In GUI settings, Self-Grounded Verification~\citep{andrade2025selfgrounded} reveals agreement bias in MLLM verifiers, and failure taxonomies~\citep{cemri2025multiagentfail} identify premature completion and action loops as dominant errors. \name directly targets these issues with evidence-grounded Completeness-Aware Planner and a multi-level Loop Breaker that escalates recovery strategies.

\section{The \name System}
\label{sec:method}
\begin{figure}[t]
    \centering
    \includegraphics[width=1\linewidth]{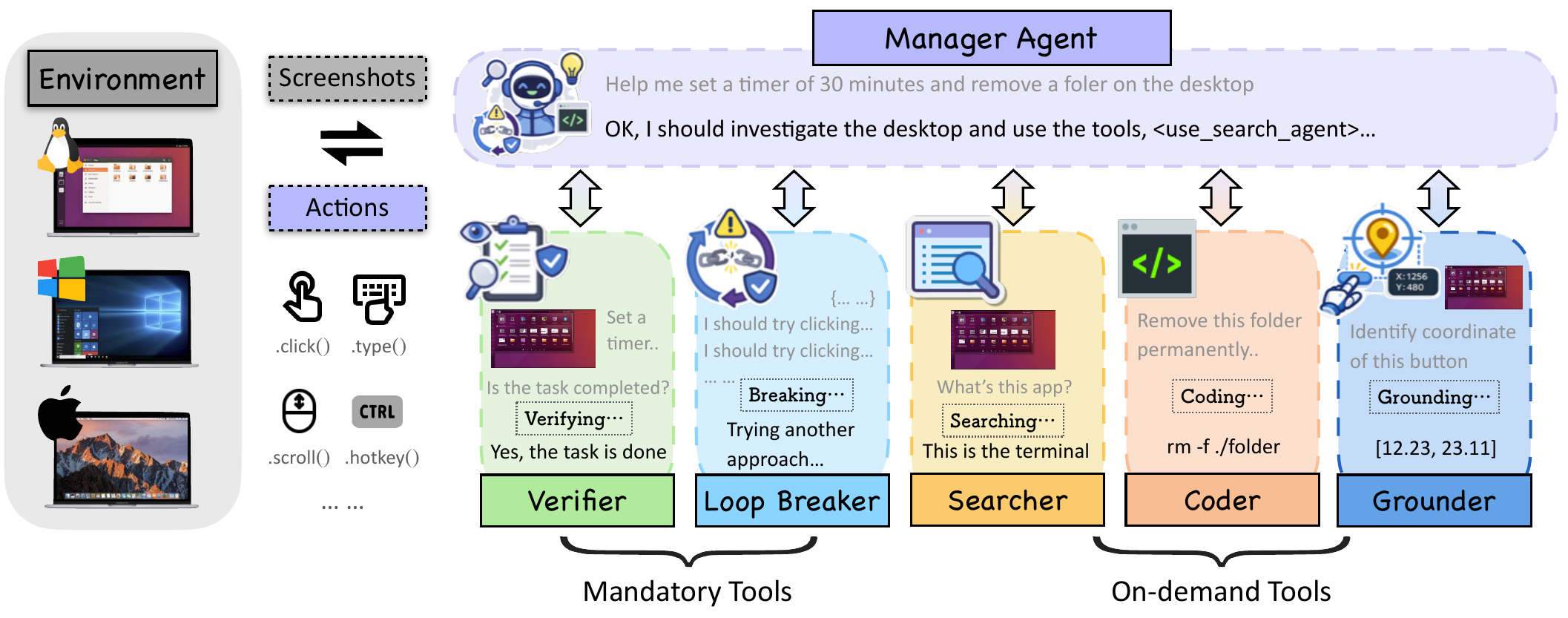}
    \caption{
    \textbf{Overview of \name.} The Manager Agent perceives screenshots and acts through basic GUI actions (\eg, \texttt{click}, \texttt{type}, \texttt{scroll}, \texttt{hotkey}). 
    Two \emph{mandatory} modules coordinate with the Manager: the Verifier guards against premature completion, and the Loop Breaker interrupts stalled action loops. 
    In the meanwhile, three \emph{on-demand} tools: Search Agent (Searcher), Coding Agent (Coder), and Grounding Agent (Grounder), are triggered by the Manager as needed.
    }
     \label{fig:pipeline}
   \end{figure}
\subsection{Overview}
\label{sec:architecture}

\name is built around a Manager agent that performs GUI interaction in a perceive–reason–act loop as shown in Figure~\ref{fig:pipeline}. At step~$t$, its belief state $b_t$ summarizes the user instruction, trajectory history, and current observation, including the screenshot and tool results. From this state, it selects an action $a_t$: a UI primitive (\eg, clicking or scrolling through \texttt{pyautogui}) or a call to the Search, Coding, or Grounding Agent. Keeping these choices in a shared action space allows the Manager to select the form of assistance appropriate to the current obstacle while retaining responsibility for the task throughout execution.

\subsection{Completeness-Aware Planner}
\label{sec:completion_aware_planning}

Conventional hierarchical planning decomposes how to execute the task~\citep{agashe2024agents,agashe2025agents2}. These designs primarily focus on how a task should be decomposed, but they provide weaker signals regarding whether the requested outcome has actually been achieved. Our \emph{Completeness-Aware Planner} (CAP) instead represents what must be true for the task to be complete. It combines a \emph{Completion Gate} for evidence checks with a separate model judge for proposed completion claims.

\noindent \textbf{Completion Gate.}
To make completion assessable as execution unfolds, the gate first derives $K$ UI-observable success criteria $\mathcal{C} = \{c_1, \ldots, c_K\}$ from the task instruction. Expressing these criteria in terms of the requested final state lets the Manager reconsider its route without losing track of what must ultimately be demonstrated. The criteria do not prescribe an action order or a dependency graph; their evidence status is revisited at each step to identify what still requires execution or verification. Hidden-state requirements are represented by observable proxies where possible, whose adequacy remains a limitation. At each step~$t$, the Manager checks the available evidence against these criteria using its belief state $b_t$:
\begin{equation}
  \text{Gate}(b_t) =
  \begin{cases}
    \textsc{done} & \text{if self-check passes } K \text{ criteria and UI is stable} \\
    \textsc{continue} & \text{otherwise}
  \end{cases}
  \label{eq:gate}
\end{equation}
Because UI changes may lag behind an action, an immediate screenshot can leave its outcome unresolved. Action-specific verification rules
 ask the Manager to check the expected visual outcome after each action (Table~\ref{tab:micro_verify} in Appendix~\ref{app:impl}), helping
  it distinguish pending updates from evidence that a requirement has been satisfied.

\noindent \textbf{Completeness Model Judge.}
Because the Completion Gate only conditions on the initial task instruction, the Manager can still miss potential task requirements, incurring premature task termination even with the Completion Gate. Once the gate outputs \textsc{done}, a separate MLLM judge therefore revisits the original instruction alongside the current observation and recent trajectory, rather than relying solely on the initial criterion list. This second perspective is intended to expose missing requirements or unsupported completion claims at the point where they would otherwise end execution.

Given the instruction $g$, observation $o_t$, and trajectory context summarized in $b_t$, the judge returns an accept/reject decision with
 an explanation. Acceptance requires direct visual evidence for every criterion, visible confirmation of side-effect actions such as saving,
  exporting, or sending, and a stable UI. Termination requires agreement between the gate and the judge. On rejection, criterion-level 
  evidence and missing-evidence feedback are appended to the trajectory, giving the Manager a basis for further action. Thus, the final
   decision is binary while the feedback preserves partial progress. The judge's instructions appear in Appendix~\ref{app:verifier}.
   
\subsection{Loop Breaker}
\label{sec:loop_breaker}

Repeated failures can reflect an ineffective interaction, an unproductive workflow, or a broader trajectory pattern. The Loop Breaker addresses these at three levels using screen observations $o_t$ and actions $a_t$. It runs after every Manager action: Tiers 1--2 check local action and screen-state recurrence, while Tier 3 invokes an external model to inspect the recent trajectory and produce a strategy-level signal.
We define counters for repeated actions without visible change and recurring screen states:
\begin{equation}
  n_t^{a} = \bigl|\{i \in [t\!-\!1,\, t] : a_i = a_t \;\land\; o_{i+1} \approx o_i\}\bigr|, \qquad
  n_t^{o} = \bigl|\{i \in [t\!-\!2,\, t] : o_i \approx o_t\}\bigr|
  \label{eq:loop_counters}
\end{equation}
These counters and an external reflection signal $w_t$ trigger mandatory intervention, with fixed thresholds
$\tau_a=2$ and $\tau_o=3$ across tasks and both benchmark platforms.

\noindent \textbf{Tier~1: Modality Switch ($n_t^{a} \geq \tau_a$).}
If the same action on the same target repeatedly produces no visible change, the agent must switch interaction mode
(\eg, keyboard shortcut $\to$ menu click $\to$ command-line).

\noindent \textbf{Tier~2: Strategy Change ($n_t^{o} \geq \tau_o$).}
If the same screen state recurs three times without progress, the agent must change its overall strategy
(\eg, from menu navigation to programmatic file editing). Repetition alone does not establish task infeasibility;
failure termination requires separate evidence.

\noindent \textbf{Tier~3: Reflection-Driven Judge.}
To detect unproductive patterns beyond local repetition, an external model judge inspects the recent trajectory
and returns a ``keep'' or ``switch'' signal. When $w_t$ signals a switch, a directive at step $t+1$ blacklists
the repeated action and requires the Manager to choose an alternative (\eg, \texttt{click} $\to$ \texttt{type},
or GUI actions $\to$ \texttt{call\_coding\_agent}). The prompts appear in Appendix~\ref{app:reflection}.

Together, the three tiers connect local action failures, recurring navigation dead ends, and broader trajectory
patterns to changes in interaction mode or strategy.

\subsection{Search Agent (Searcher)}
\label{sec:search}
Recovery can change the route, but an unfamiliar application may leave the agent without a useful alternative to try. A screenshot reveals the current interface while often leaving the procedure for reaching a goal implicit. In such cases, a short tutorial can supply the missing workflow knowledge and give subsequent interaction a more informed starting point.

Retrieving that knowledge through visual browser search, as in OS-Symphony~\citep{yang2026ossymphony}, introduces additional navigation and grounding demands. Since the information sought is procedural text, we expose a direct textual interface through the Search Agent $\mathcal{S}$. The Manager invokes \texttt{search(query)} within the same action space as UI primitives, formulating a targeted ``How to'' question according to guidelines in the tool description. An LLM with native search grounding (\eg, Gemini~3~Pro~\citep{gemini3pro2025}) returns a summarized tutorial as plain text, which is added to the belief state for subsequent steps. This avoids browser interaction for retrieval while leaving the Manager to apply the guidance in the current environment.

\texttt{Search} is useful when uncertainty concerns a workflow for which external guidance is likely to exist. Accordingly, the tool is invoked on demand when the Manager is uncertain about a GUI workflow and a tutorial is likely to be available, as with well-documented application features.

In addition to the above components, we also equip the Manager agent with a grounding agent (Grounder) for screen elements localization and a coding agent (Coder) for code executions~\citep{gonzalez2025agents3, yang2026ossymphony, hippo2026}.

\section{Experiments}
\subsection{Experimental Settings}
\noindent \textbf{Benchmarks.} We evaluate our \name on two desktop GUI benchmarks. OSWorld-Verified (OSWorld for short)~\citep{xie2024osworld} is our primary benchmark. Following common practice~\citep{xie2024osworld}, we use the 361-task evaluation set that excludes the eight Google Drive tasks, spanning five application domains (OS, Office, Daily, Professional, and Workflow) in an Ubuntu virtual environment. 
WindowsAgentArena~(WAA)~\citep{bonatti2024waa} provides 154 tasks on a Windows environment spanning productivity, file management, and 
system configuration, serving as our cross-platform generalization testbed.

\begin{table}[t]
\centering
\caption{Success rates (\%) on OSWorld-Verified~\citep{xie2024osworld}. Our results use the 361-task evaluation set. We report results of 15, 50, and 100 steps budgets. Baseline results are taken from the OSWorld leaderboard, except OS-Symphony at 100 steps, which is taken from its paper.}
\scriptsize
\begin{tabular}{l c cccccc}
\toprule
\textbf{Method} & \textbf{Step} &
\multicolumn{6}{c}{\textbf{Success Rate (\%)}} \\
\cmidrule(lr){3-8}
 & & OS & Office & Daily & Prof. & Work. & Avg. \\
\midrule
\multicolumn{8}{c}{\textit{Max 15 Steps}} \\
\midrule
\name w/ Gemini 3 Flash      & 15 & 79.20 & 29.92 & 54.14 & 57.13 & 34.00 & 43.15 \\
\name w/ Gemini 3.1 Pro      & 15 & 83.30 & 52.99 & 54.33 & 61.22 & 34.70 & 51.69 \\
\name w/ Sonnet 4.6      & 15 & 83.30 & \textbf{69.72} & 58.72 & 57.13 & \textbf{60.20} & 64.13 \\
\name w/ Opus 4.5      & 15 & \textbf{87.50} & 57.16 & 59.69 & 71.43 & 45.20 & 58.58 \\
\name w/ Opus 4.6      & 15 & 83.30 & 60.65 & \textbf{66.38} & \textbf{79.59} & 55.90 & \textbf{64.75} \\
\midrule
\multicolumn{8}{c}{\textit{Max 50 Steps}} \\
\midrule
UiPath w/ GPT-5~\citep{uipath2025}        & 50  & 73.91 & 49.52 & 62.12 & 71.43 & 37.30 & 53.69 \\
CoAct-1 w/ GPT-5~\citep{song2025coact}       & 50  & 70.83 & 60.65 & 54.09 & 69.39 & 42.37 & 56.39 \\
OS-Symphony w/ GPT-5~\citep{yang2026ossymphony}   & 50 & 78.26 & 65.74 & 65.92 & 79.59 & 55.30 & 65.77 \\
\midrule
\name w/ Gemini 3 Flash      & 50 & 83.30 & 69.19 & 62.75 & 62.49 & 51.00 & 63.14 \\
\name w/ Gemini 3.1 Pro      & 50 & 83.30 & 71.51 & 67.32 & 67.35 & 55.90 & 66.80 \\
\name w/ Sonnet 4.6      & 50 & 83.30 & \textbf{79.23} & 69.24 & 57.13 & \textbf{66.70} & 71.11 \\
\name w/ Opus 4.5      & 50 & \textbf{87.50} & 75.97 & 71.38 & 79.59 & 55.60 & 71.00 \\
\name w/ Opus 4.6      & 50 & 83.30 & 78.60 & \textbf{72.60} & \textbf{83.67} & 61.30 & \textbf{73.85} \\

\midrule
\multicolumn{8}{c}{\textit{Max 100 Steps}} \\
\midrule
Seed-1.8 \citep{seed18}          & 100 & 66.67 & 68.80 & 67.05 & 71.43 & 42.38 & 61.87 \\
Claude-Sonnet-4.5 \citep{claudesonnet45}    & 100 & 70.83 & 72.59 & 61.35 & 63.27 & 49.54 & 62.84 \\
Kimi K2.6 \citep{kimiteam2026kimik26}    & 100 & 79.17 & 80.03 & 77.11 & 81.63 & 54.97 & 73.07 \\

CoAct-1 w/ GPT-5  \citep{song2025coact}     & 100 & 75.00 & 62.93 & 57.94 & 71.43 & 47.87 & 59.93 \\
GTA1 w/ GPT-5  \citep{yang2025gta1}       & 100 & 79.17 & 63.91 & 62.56 & 79.59 & 50.91 & 63.41 \\
OS-Symphony w/ GPT-5   \citep{yang2026ossymphony}             & 100 & 79.17 & 65.73 & 67.76 & 69.23 & 57.98 & 65.84 \\
Agent S3 w/ GPT-5 \citep{gonzalez2025agents3}     & 100 & 77.50 & 66.30 & 61.36 & 69.80 & 51.37 & 62.60 \\

UiPath w/ Opus 4.5~\citep{uipath2025}             & 100 & 70.83 & 74.13 & 68.33 & 73.47 & 52.97 & 67.14 \\
Agent S3 w/ Opus 4.5  \citep{gonzalez2025agents3}    & 100 & 75.00 & 75.53 & 63.94 & 58.37 & 57.43 & 66.00 \\
HIPPO w/ Opus 4.5~\citep{hippo2026}             & 100 & 87.50 & 74.27 & 69.27 & \textbf{95.92} & 64.31 & 74.49 \\
\midrule
\name w/ Gemini 3 Flash      & 100 & 91.70 & 64.90 & 74.85 & 67.33 & 63.40 & 68.77 \\
\name w/ Gemini 3.1 Pro      & 100 & 83.30 & 76.60 & 73.76 & 73.47 & 62.90 & 72.47 \\
\name w/ Sonnet 4.6      & 100 & 83.30 & 79.24 & 69.25 & 57.14 & \textbf{68.80} & 71.67 \\
\name w/ Opus 4.5      & 100 & 87.50 & 79.38 & \textbf{76.57} & 81.63 & 61.00 & 74.89 \\
\name w/ Opus 4.5 + MAI-UI  & 100 & 91.67 & \textbf{84.26} & 72.77 & 83.67 & 61.09 & 76.26 \\
\name w/ Opus 4.6      & 100 & \textbf{91.70} & 82.87 & 75.17 & 83.67 & 65.60 & \textbf{77.45} \\

\bottomrule
\end{tabular}
\label{tab:main_results}
\end{table}

\noindent \textbf{Implementation Details.} In each configuration, the named backbone serves as the Manager, Reflection Agent, Completeness Model Judge, and Coding Agent, with Gemini~3~Pro~\citep{gemini3pro2025} or Gemini~3.1~Pro~\citep{gemini31pro2025} as the Search Agent. Seed~1.8~\citep{seed18} is the default grounding model, except for one Opus~4.5 variant that uses MAI-UI~\citep{zhou2025maiuitechnicalreportrealworld}; on WAA, Gemini~3~Flash is the Manager.
The Completeness Model Judge within Completeness-Aware Planner uses temperature $T\!=\!0.2$ and all other components use $T\!=\!1.0$. Full model assignments and infrastructure are given in Appendix~\ref{app:impl}.

\subsection{Main Results}
\paragraph{\name Surpasses Human-level Performance.}
With a 100-step budget, \name with Opus~4.5 achieves 74.89\% and with Opus~4.6 achieves 77.45\%, exceeding the reported human reference level of 72.4\%.
Notably, three of the five backbones surpass human performance at 100 steps: Opus~4.6 (77.45\%), Opus~4.5 (74.89\%), and Gemini~3.1~Pro (72.47\%), demonstrating that \name's framework design generalizes across model families.
When varying the Manager Agent backbone, \name w/ Sonnet~4.6 achieves an overall 71.67\%, with particularly strong results in the OS and Multi-Apps domains (per OSWorld's official task categorization~\citep{xie2024osworld}), where it surpasses Agent~S3 w/ Opus~4.5 by 8.3\% and 11.4\%, respectively (\ie average 76\% \vs 66\%).
Moreover, even Gemini~3~Flash achieves 68.77\%, outperforming the best GPT-5-based frameworks despite using a substantially lighter model, providing additional evidence that our \name can effectively adapt to models with different capabilities.
\begin{wraptable}{r}{0.52\columnwidth}
\vspace{-1.5em}
\centering
\caption{Ablation of components on OSWorld. Component effects depend on the backbone and budget.}
\small
\setlength{\tabcolsep}{4pt}
\begin{tabular}{lcc}
\toprule
\textbf{Method} & \multicolumn{2}{c}{\textbf{Success Rate (\%)}} \\
\cmidrule(lr){2-3}
 & \textbf{50 Steps} & \textbf{100 Steps} \\
\midrule
\name w/ Gemini 3 Flash & 63.14 & \textbf{68.77} \\
\quad $-$ CAP & \textbf{66.00} & 67.34 \\
\quad $-$ Loop Breaker & 58.90 & 66.95 \\
\quad $-$ Search Agent & 62.54 & 65.82 \\
\midrule
\name w/ Sonnet 4.6 & \textbf{71.11} & \textbf{71.67} \\
\quad $-$ CAP & 68.53 & 68.81  \\
\quad $-$ Loop Breaker & 69.67 & 71.63  \\
\quad $-$ Search Agent & 68.92 & 70.04  \\
\bottomrule
\end{tabular}
\label{tab:components}
\vspace{-1.5em}
\end{wraptable}
Across categories, Opus~4.6 and Gemini~3~Flash both exceed 91\% on the OS domain at 100 steps, outperforming the strongest baseline (HIPPO at 87.50\%) by 4.2\% and other baselines by over 12\% (\eg, OS-Symphony at 79.17\%), indicating reliable system-level task handling across backbones.
The Professional domain shows the largest variance (57.1\% Sonnet~4.6 vs.\ 83.7\% Opus~4.6), suggesting that the reasoning depth of the Manager matters more in difficult tasks.

Additionally, \name generalizes well to WindowsAgentArena (Table~\ref{tab:waa_main}). On WAA, \name reaches 60.4\% overall at 50 steps and 61.0\% at 100 steps, which outperforms Agent~S3 w/ GPT-5 at both configurations by at least 4.4\% (\ie, Agent S3 56.6\% at 100 steps), and GTA1 w/ o3 at 100 steps by 9.8\%, providing evidence of \name's ability to generalize across operating systems and task distributions.

\begin{table}[t]
\centering
\caption{Results on WindowsAgentArena (WAA)~\citep{bonatti2024waa}. Our \name (Gemini 3 Flash) achieves the best performance across both 50 and 100 step configurations.}
\small
\setlength{\tabcolsep}{3pt}
\begin{tabular}{l ccccccc}
\toprule
\textbf{Method} & \textbf{Office} & \textbf{Web} & \textbf{Sys.} & \textbf{Code} & \textbf{Media} & \textbf{Util.} & \textbf{Overall} \\
\midrule

\multicolumn{8}{c}{\textit{Max 50 Steps}} \\
\midrule
Qwen3-VL-32B \citep{bai2025qwen3vl}   & 19.1 & 49.7 & 54.2 & 21.1 & 42.2 & 25.0 & 31.7 \\
UI-TARS-2 \citep{wang2025uitars2technicalreportadvancing}   & - & - & - & - & - & - & 50.6 \\
Agent S3 w/ GPT-5~\citep{gonzalez2025agents3}    & - & - & - & - & - & - & 54.1 \\
\midrule
\name          & 32.6 & 73.3 & 87.5 & 66.7 & 52.4 & 75.0 & \textbf{60.4} \\
\quad $-$ Completeness-Aware Planner      & 11.6 & 73.3 & 70.8 & 54.2 & 61.9 & 83.3 & 51.3 \\
\quad $-$ Loop Breaker      & 27.9 & 66.7 & 83.3 & 58.3 & 38.1 & 75.0 & 52.6 \\
\quad $-$ Search Agent    & 18.6 & 63.3 & 75.0 & 66.7 & 38.1 & 66.7 & 49.4 \\

\midrule
\multicolumn{8}{c}{\textit{Max 100 Steps}} \\
\midrule
GTA1-32B w/ o3~\citep{yang2025gta1}    & - & - & - & - & - & - & 51.2 \\
Agent S3 w/ GPT-5~\citep{gonzalez2025agents3}    & - & - & - & - & - & - & 56.6 \\
\midrule
\name          & 35.0 & 73.3 & 87.5 & 66.7 & 52.4 & 83.3 & \textbf{61.0} \\
\quad $-$ Completeness-Aware Planner      & 13.9 & 73.3 & 70.8 & 54.2 & 57.1 & 83.3 & 51.3 \\
\quad $-$ Loop Breaker      & 30.2 & 66.7 & 83.3 & 58.3 & 47.6 & 83.3 & 55.8 \\
\quad $-$ Search Agent    & 25.0 & 63.3 & 75.0 & 66.7 & 47.6 & 83.3 & 53.9 \\

\bottomrule
\end{tabular}
\vspace{-1em}
\label{tab:waa_main}
\end{table}

\paragraph{Configuration-Controlled Comparisons.}
\label{sec:matched_comparisons}
Under our harness, the Sonnet~4.6 base scaffold (w/ coder + grounder) scores 69.19\% on the 361 tasks, compared with 71.67\% for the full framework, a 2.48\% gain with the same backbone, grounding, action interface, and budget.\footnote{The public Sonnet~4.6 result of 72.1\%~\citep{claudesonnet46} uses Anthropic's native computer-use tools and screen-cropping grounding and is therefore a system-level reference rather than a control.}
With Opus~4.5, \name achieves 74.89\% using the Seed~1.8 grounder, already exceeding HIPPO's 74.49\% with the stronger MAI-UI grounder~\citep{zhou2025maiuitechnicalreportrealworld}. Replacing only our grounder with MAI-UI raises \name's score to 76.26\%, widening the advantage over HIPPO to 1.77\% under the same backbone and grounder.

\subsection{Verify Each Component}
To understand the effect of each component in \name, we conduct an ablation study and analyze early stopping and loop behavior. Table~\ref{tab:components} reports the results across different MLLM backbones and action budgets on OSWorld.
For Sonnet~4.6 at 100 steps, removing Completeness-Aware Planner lowers the score by 2.86\% (71.67\% to 68.81\%), while removing search lowers it by 1.63\% and removing loop control by only 0.04\%.
For Gemini~3~Flash, loop control gives a 4.24\% gain at 50 steps and search a 2.95\% gain at 100 steps.
Verification is not uniformly beneficial: at 50 steps, Gemini~3~Flash scores 66.00\% without it, compared with 63.14\% in the full system, a 2.86\% decrease when verification is included. At 100 steps its contribution is positive (68.77\% versus 67.34\%).
The results support configuration-dependent benefits rather than a claim that every module always improves performance.

On WAA (Table~\ref{tab:waa_main}), removing Search, verification, and Loop Breaker lowers the 50-step score by 11.0\%, 9.1\%, and 7.8\%, respectively.
At 100 steps the corresponding drops are 7.1\%, 9.7\%, and 5.2\%.
Removing verification lowers Office accuracy from 32.6\% to 11.6\% at 50 steps, and removing search lowers it to 18.6\%.
These stronger effects on WAA complement the smaller or occasionally negative effects on OSWorld.

\begin{figure}[t]
    \centering
    \includegraphics[width=1\linewidth]{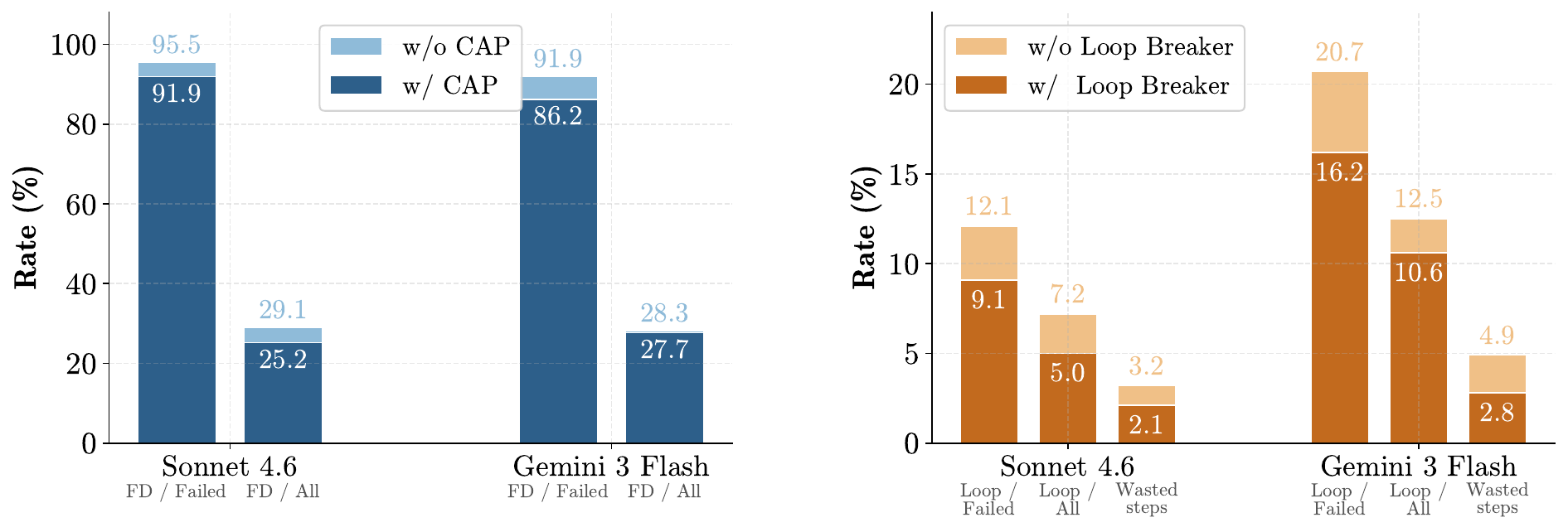}
     \caption{Component analysis on OSWorld at 100 steps. CAP reduces false completions (FD: false DONE), using Table~\ref{tab:verifier_breakdown}; All denotes all 361 tasks. The Loop Breaker reduces loop incidence and the wasted steps ratio.}
     \label{fig:component_ablation0}
     \vspace{-1em}
\end{figure}
   
\paragraph{Completeness-Aware Planner Mitigates False Completion Ratio.}
Figure~\ref{fig:component_ablation0} analyzes false completions using the 100-step results in Table~\ref{tab:verifier_breakdown}. We report \textit{False DONE/Failed} (FDF), the fraction of failed tasks with false completion, and \textit{False DONE/All} (FDA), the false DONE count divided by all 361 tasks. FDA differs from the complement of DONE accuracy, whose denominator is completion signals.
For Sonnet~4.6, CAP reduces false DONE counts from 105 to 91, lowering FDF by 3.6\% (95.5\% to 91.9\%) and FDA by 3.9\% (29.1\% to 25.2\%).
For Gemini~3~Flash, false DONE counts decrease from 102 to 100: FDF drops by 5.7\% (91.9\% to 86.2\%), while FDA decreases by 0.6\% (28.3\% to 27.7\%).
Even with CAP, FDF remains above 86\% for both backbones: most failed tasks still end with an incorrect completion claim. CAP therefore only partially resolves false completion.
\paragraph{Loop Breaker Reduces Loops and Wasted Steps.}
We similarly analyze loop behavior with three metrics: \textit{Loop / Failed} (LF) (fraction of failed tasks involving loops), \textit{Loop / All} (LA) (fraction of all tasks with loops), and \textit{Wasted Steps Ratio} (WSR) (fraction of total steps spent in detected loops) with results also in Figure~\ref{fig:component_ablation0}.
The Loop Breaker reduces loop incidence across both backbones. In detail, for Sonnet~4.6, LF drops by 3\% from 12.1\%, with wasted steps decreasing from 3.2\% to 2.1\% over all steps. 
The effect is more pronounced for Gemini~3~Flash, where LF drops by 4.5\% (from 20.7\% to 16.2\%) and wasted steps nearly halve from 4.9\% to 2.8\%.
This backbone-dependent pattern is consistent with the ablation findings: Gemini~3~Flash loops roughly twice as often as Sonnet~4.6 (\ie, 10.6\% vs.\ 5.0\% on LA), giving evidence that weaker models are more prone to repetitive behavior and benefit more from the loop detection.

\paragraph{Transfer to CoAct-1.}
\label{sec:coact_transfer}
\begin{figure}[h]
\centering
\includegraphics[width=0.9\linewidth]{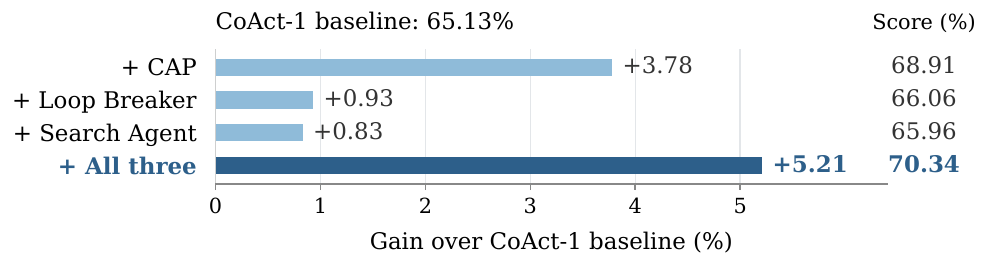}
\caption{Gains over CoAct-1 (65.13\%) on an OSWorld subset with Gemini~3~Flash and a 100-step Manager budget. CAP denotes Completeness-Aware Planner. Each proposed component helps improve CoAct-1's performance, while adding all three yields the highest score.}
\label{fig:coact_transfer}
\end{figure}
To understand whether the proposed completion, loop-recovery, and search mechanisms can improve other \emph{off-the-shelf} GUI agents, we integrate these three proposed modules into CoAct-1~\citep{song2025coact} while keeping its other components untouched.
We compare the CoAct-1 baseline with each module added individually and with all three added together on a fixed 60-task OSWorld subset. In detail, we use Gemini~3~Flash for the Manager and the CAP backbone, Seed~1.8 for grounding, and a 100-step Manager budget across all experiments; the Coder and Searcher are Gemini~3.1~Pro. We present more details in Appendix~\ref{sec:controlled_studies}.
As shown in Figure~\ref{fig:coact_transfer}, the CAP gives the largest individual score increase (65.13\% \vs 68.91\%), followed by the Loop Breaker (65.13\% to 66.06\%) and Search Agent (65.13\% to 65.96\%). Combining all three raises the score from 65.13\% to 70.34\% (+5.21\%). 
These results provide evidence that the benefits of proposed modules extend beyond our specific Manager implementation and can serve as truly portable enhancements to other GUI agents.  
Additionally, the fact that our CAP yields the largest individual gain further validates its effectiveness in addressing early stopping issues, which contributes most to performance gains in even more general GUI scenarios.

\begin{figure}[t]
    \centering
    \includegraphics[width=.8\linewidth]{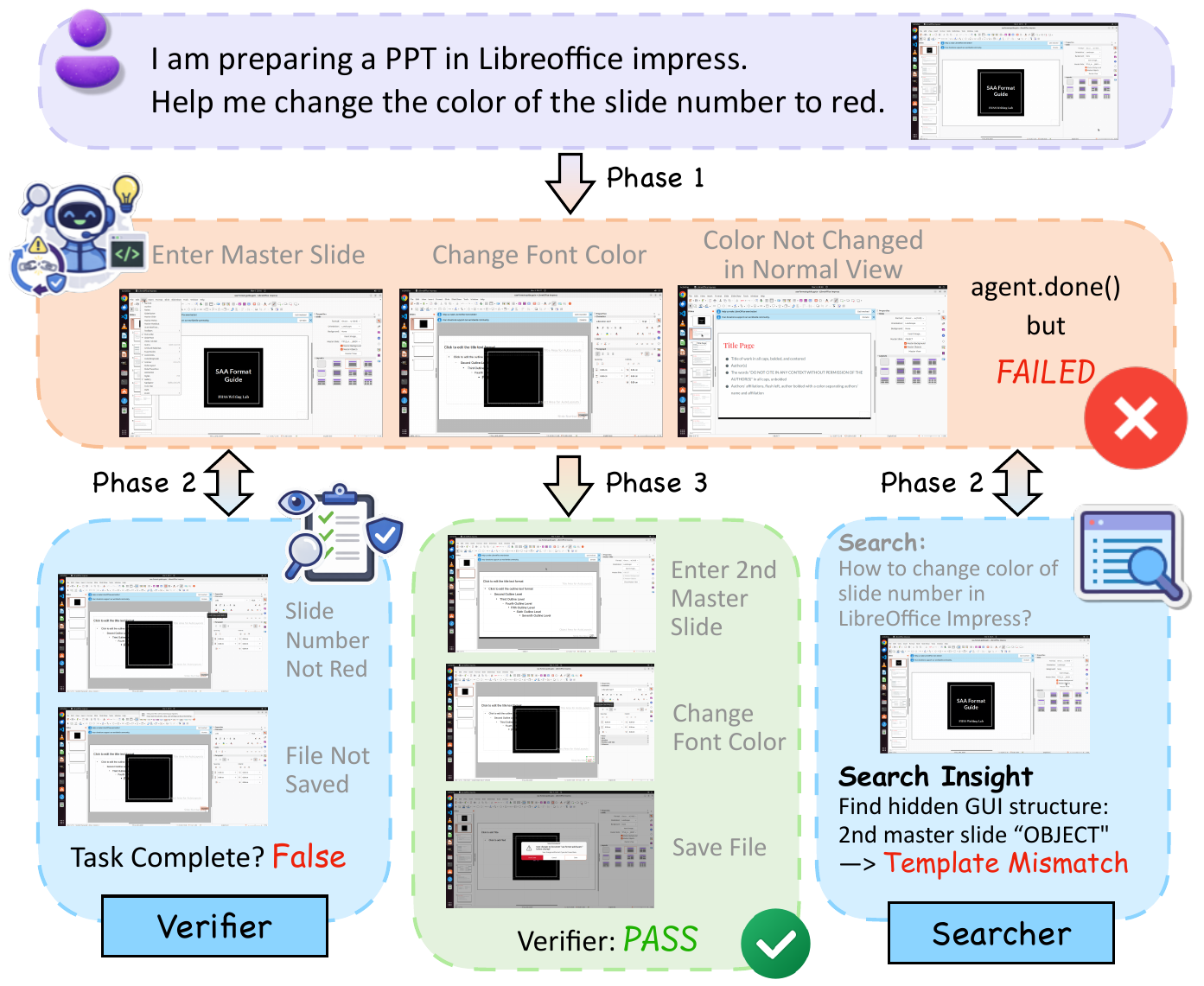}
     \caption{
     \textbf{Case study on OSWorld: \textit{Changing slide number color to red in LibreOffice Impress.}} The Manager encounters multiple failures before final completion. CAP rejects two premature \texttt{done()} calls, and the Searcher provides critical knowledge about multi-master-slide editing.}
     \label{fig:case_study}
     \vspace{-1em}
\end{figure}

\subsection{Case Study}
\label{sec:case_study}

To illustrate how components interact to recover from failures, we display a case study from OSWorld in Figure~\ref{fig:case_study}. 
The task instruction is \textit{``The slide number is barely visible. Please change the color of the slide number to red.''}.
The Manager's first attempt modifies one master slide and calls \texttt{done()}, but our CAP rejects it, noting the slide number is still gray and the file has not been saved. 
Prompted by the rejection, the Manager invokes the Searcher. The retrieved procedural knowledge contains a key insight: \textit{``If your presentation uses different master slides for different sections, repeat for each master slide.''} Acting on this, the Manager inspects the slide content and discovers a second master named \texttt{OBJECT} that it had overlooked.
With this knowledge, the Manager applies the color change to both slides and calls \texttt{done()} a second time. The CAP rejects it again: the color is now correct, but the file is still unsaved. Only after an explicit \texttt{Ctrl+S} does the Judge accept the result.
This case highlights the complementary roles: CAP prevents early termination while the Searcher supplies task-specific knowledge.

\section{Conclusion}

In this paper, we presented \name, a modular GUI agent framework that addresses premature task completion and repetitive action loops in autonomous desktop agents. Its Completeness-Aware Planner, Loop Breaker, and Search Agent push OSWorld performance to 77.5\% with Opus~4.6 and reach 61.0\% on WindowsAgentArena. Ablations show that each component contributes, with gains shaped by backbone strength and step budget. False completion remains the dominant failure mode, motivating continued research on completion verification.

\bibliographystyle{iclr2027_conference}
\bibliography{main}

\newpage
\appendix

\section{Appendix}
This supplementary material provides more implementation details, prompt templates, the full action space, and additional ablation results for \name. We organize the content as follows:
\begin{itemize}
    \item Section~\ref{sec:controlled_studies}: Controlled component and transfer studies, including the complete 60-task manifest in Section~\ref{app:task_manifest}.
    \item Section~\ref{app:subset}: Composition of the fixed 60-task subset.
    \item Section~\ref{sec:inference_cost}: Inference cost and tool attribution.
    \item Section~\ref{app:limitations}: Broader impact and limitations.
    \item Section~\ref{app:impl}: Implementation details including model configurations and benchmark categories.
    \item Section~\ref{app:action_space}: The full action space exposed to the Manager Agent.
    \item Section~\ref{app:prompts}: The system prompts for different agent roles.
    \item Section~\ref{app:ablations}: Additional ablation results.
\end{itemize}

\subsection{Controlled Component and Transfer Studies}
\label{sec:controlled_studies}

\paragraph{Setup.}
We supplement the main evaluation with component, transfer, and usage studies on a fixed, non-random 60-task subset selected from the
 same 361-task OSWorld-Verified evaluation set used for the main experiments.
The subset was fixed before the controlled experiments and selected to approximately preserve application-level coverage, without reference
 to the outcomes of the evaluated module configurations. Its composition is given in Table~\ref{tab:subset_coverage}, and the complete task
 manifest is provided in Appendix~\ref{app:task_manifest}.
All subset configurations use the same task IDs, Gemini~3~Flash for the Manager and Reflection backbone, Seed~1.8 grounding, the same GUI action space, and a 100-step Manager budget.
We evaluate this fixed subset only with the 100-step configuration. Every subset score in the CoAct-1 transfer experiment (Figure~\ref{fig:coact_transfer}), and usage analysis (Table~\ref{tab:usage_subset}) is obtained under this budget, rather than averaged across step budgets. No 15- or 50-step configurations are evaluated for these subset studies.
In these additional studies, the Coder and Searcher use Gemini~3.1~Pro when present, which differs from the main-suite Gemini~3~Flash configuration in Table~\ref{tab:model_config}.
The base scaffold (w/ coder + grounder) retains standard coding and grounding access; the three experimental factors are completion verification, loop control, and search.
We report aggregate scores as subset point estimates, without confidence intervals or significance claims.
All differences below are calculated by subtracting the displayed percentage scores and are denoted by \%.

\paragraph{Recovery thresholds and scope of generalization.}
The same two-attempt action trigger and three-occurrence screen trigger are used across tasks; they are not tuned to individual OSWorld scenarios.
On WAA, removing the same Loop Breaker design reduces scores by 7.8\% at 50 steps (60.4\% to 52.6\%) and by 5.2\% at 100 steps (61.0\% to 55.8\%).
Together with the CoAct-1 transfer, this supports the usefulness of the recovery design across the tested platforms and scaffolds.
It does not establish threshold optimality or robustness to arbitrary distributions: we have not performed a threshold-sensitivity sweep, and the negligible 0.04\% Sonnet~4.6 gain at 100 steps illustrates that explicit recovery is most useful when the agent is prone to loops.

\subsubsection{Task IDs and Descriptions}
\label{app:task_manifest}
Table~\ref{tab:task_manifest} lists all 60 task IDs used in the fixed subset, grouped by application domain, with concise English descriptions of the tasks.
The domain counts are Chrome (8), GIMP (4), LibreOffice Calc (8), LibreOffice Impress (8), LibreOffice Writer (4), Multi-app (15), OS (4), Thunderbird (3), VLC (3), and VS Code (3).
The coverage percentages are reported in Table~\ref{tab:subset_coverage}.
A dagger ($\dagger$) preserves the five tasks flagged in the subset specification for nonexistent or unsupported capabilities, or restrictive tool constraints. This annotation describes the task specification, not an observed success or failure of the evaluated agent.

\begingroup
\footnotesize
\setlength{\LTleft}{0pt}
\setlength{\LTright}{0pt}
\renewcommand{\arraystretch}{1.12}
\begin{longtable}{@{}>{\raggedright\arraybackslash}p{0.53\textwidth}>{\raggedright\arraybackslash}p{\dimexpr0.47\textwidth-2\tabcolsep\relax}@{}}
\caption{Task manifest for the fixed 60-task OSWorld-Verified subset. Descriptions summarize the task specifications; $\dagger$ marks the five flagged cases described above.}
\label{tab:task_manifest}\\
\toprule
\textbf{Task ID} & \textbf{Task description} \\
\midrule
\endfirsthead
\multicolumn{2}{@{}l}{\tablename~\thetable\ (continued)}\\
\toprule
\textbf{Task ID} & \textbf{Task description} \\
\midrule
\endhead
\midrule
\multicolumn{2}{r@{}}{\textit{Continued on next page}}\\
\endfoot
\bottomrule
\endlastfoot
\multicolumn{2}{@{}l}{\textbf{Chrome (8)}}\\*
\texttt{bb5e4c0d-f964-439c-97b6-bdb9747de3f4} & Set Bing as the default search engine. \\
\texttt{af630914-714e-4a24-a7bb-f9af687d3b91} & Increase font size in accessibility settings. \\
\texttt{99146c54-4f37-4ab8-9327-5f3291665e1e} & Automatically clear browsing data on exit. \\
\texttt{6766f2b8-8a72-417f-a9e5-56fcaa735837} & Extract and locally install an extension. \\
\texttt{3720f614-37fd-4d04-8a6b-76f54f8c222d} & $\dagger$ Change to a nonexistent language. \\
\texttt{fc6d8143-9452-4171-9459-7f515143419a} & Look up flights through live web interaction. \\
\texttt{7f52cab9-535c-4835-ac8c-391ee64dc930} & Search and filter results in Google Shopping. \\
\texttt{f0b971a1-6831-4b9b-a50e-22a6e47f45ba} & Retrieve a Super Bowl score from the web. \\
\addlinespace
\multicolumn{2}{@{}l}{\textbf{GIMP (4)}}\\*
\texttt{7a4deb26-d57d-4ea9-9a73-630f66a7b568} & Reduce brightness using a filter. \\
\texttt{2a729ded-3296-423d-aec4-7dd55ed5fbb3} & Make the background transparent using a selection. \\
\texttt{d16c99dc-2a1e-46f2-b350-d97c86c85c15} & Resize a layer. \\
\texttt{045bf3ff-9077-4b86-b483-a1040a949cff} & $\dagger$ Convert to CMYK mode, which GIMP does not support in the task setting. \\
\addlinespace
\multicolumn{2}{@{}l}{\textbf{LibreOffice Calc (8)}}\\*
\texttt{357ef137-7eeb-4c80-a3bb-0951f26a8aff} & Compute work hours times hourly wages with formulas. \\
\texttt{1954cced-e748-45c4-9c26-9855b97fbc5e} & Create a pivot table. \\
\texttt{12382c62-0cd1-4bf2-bdc8-1d20bf9b2371} & Create a column chart. \\
\texttt{37608790-6147-45d0-9f20-1137bb35703d} & Split a name column for data cleaning. \\
\texttt{3aaa4e37-dc91-482e-99af-132a612d40f3} & Export a CSV file. \\
\texttt{0bf05a7d-b28b-44d2-955a-50b41e24012a} & Pad IDs with leading zeros. \\
\texttt{4188d3a4-077d-46b7-9c86-23e1a036f6c1} & Freeze the first row. \\
\texttt{8b1ce5f2-59d2-4dcc-b0b0-666a714b9a14} & Highlight weekends with conditional formatting. \\
\addlinespace
\multicolumn{2}{@{}l}{\textbf{LibreOffice Impress (8)}}\\*
\texttt{5d901039-a89c-4bfb-967b-bf66f4df075e} & Turn an image into a cover slide (complex editing). \\
\texttt{3b27600c-3668-4abd-8f84-7bcdebbccbdb} & Set a blue background on all slides. \\
\texttt{39be0d19-634d-4475-8768-09c130f5425d} & Insert a table. \\
\texttt{9ec204e4-f0a3-42f8-8458-b772a6797cab} & Duplicate slides and interleave them (difficult editing). \\
\texttt{986fc832-6af2-417c-8845-9272b3a1528b} & Underline and change the font of body text only. \\
\texttt{0a211154-fda0-48d0-9274-eaac4ce5486d} & Set yellow backgrounds only on slides containing photos of people. \\
\texttt{455d3c66-7dc6-4537-a39a-36d3e9119df7} & Export slides as PNG images. \\
\texttt{0f84bef9-9790-432e-92b7-eece357603fb} & Disable dual-screen presentation mode. \\
\addlinespace
\multicolumn{2}{@{}l}{\textbf{LibreOffice Writer (4)}}\\*
\texttt{0e47de2a-32e0-456c-a366-8c607ef7a9d2} & Add a page number on the left side of the footer. \\
\texttt{0b17a146-2934-46c7-8727-73ff6b6483e8} & Format the numeral in H$_2$O as a subscript. \\
\texttt{66399b0d-8fda-4618-95c4-bfc6191617e9} & Insert a $7\times5$ table. \\
\texttt{4bcb1253-a636-4df4-8cb0-a35c04dfef31} & Export a PDF file. \\
\addlinespace
\multicolumn{2}{@{}l}{\textbf{Multi-app (15)}}\\*
\texttt{2c9fc0de-3ee7-45e1-a5df-c86206ad78b5} & Run \texttt{git push} in a terminal. \\
\texttt{f918266a-b3e0-4914-865d-4faa564f1aef} & Run a Python script and obtain its output. \\
\texttt{f7dfbef3-7697-431c-883a-db8583a4e4f9} & Batch-convert DOC files to PDF from the command line. \\
\texttt{81c425f5-78f3-4771-afd6-3d2973825947} & Transfer data from Calc to Impress. \\
\texttt{58565672-7bfe-48ab-b828-db349231de6b} & Open an AWS billing link from an email. \\
\texttt{c2751594-0cd5-4088-be1b-b5f2f9ec97c4} & Export an image from an email attachment and edit it in GIMP. \\
\texttt{47f7c0ce-a5fb-4100-a5e6-65cd0e7429e5} & Extract a video frame and set it as the wallpaper. \\
\texttt{2fe4b718-3bd7-46ec-bdce-b184f5653624} & Create an animation using VLC and GIMP. \\
\texttt{67890eb6-6ce5-4c00-9e3d-fb4972699b06} & Find ACL best papers and enter them in a spreadsheet. \\
\texttt{6f4073b8-d8ea-4ade-8a18-c5d1d5d5aa9a} & Compile statistics on conference host cities. \\
\texttt{bc2b57f3-686d-4ec9-87ce-edf850b7e442} & Complete a data-analysis assignment following \texttt{reminder.docx}. \\
\texttt{869de13e-bef9-4b91-ba51-f6708c40b096} & Sort and organize desktop files. \\
\texttt{26150609-0da3-4a7d-8868-0faf9c5f01bb} & Complete a Snake game coding task. \\
\texttt{2373b66a-092d-44cb-bfd7-82e86e7a3b4d} & Monitor system resources with \texttt{sar}. \\
\texttt{6d72aad6-187a-4392-a4c4-ed87269c51cf} & $\dagger$ Convert a presentation to video using only Impress. \\
\addlinespace
\multicolumn{2}{@{}l}{\textbf{OS (4)}}\\*
\texttt{94d95f96-9699-4208-98ba-3c3119edf9c2} & Install Spotify. \\
\texttt{f9be0997-4b7c-45c5-b05c-4612b44a6118} & Turn off Do Not Disturb in notification settings. \\
\texttt{5812b315-e7bd-4265-b51f-863c02174c28} & Create a restricted SSH user (advanced administration). \\
\texttt{c288e301-e626-4b98-a1ab-159dcb162af5} & $\dagger$ Set the default interpreter to nonexistent Python 4. \\
\addlinespace
\multicolumn{2}{@{}l}{\textbf{Thunderbird (3)}}\\*
\texttt{15c3b339-88f7-4a86-ab16-e71c58dcb01e} & Configure an Outlook account. \\
\texttt{5203d847-2572-4150-912a-03f062254390} & Create a local folder and automatic filtering rules. \\
\texttt{d38192b0-17dc-4e1d-99c3-786d0117de77} & Compose an email and add an attachment. \\
\addlinespace
\multicolumn{2}{@{}l}{\textbf{VLC (3)}}\\*
\texttt{8f080098-ddb1-424c-b438-4e96e5e4786e} & Convert a video to MP3. \\
\texttt{aa4b5023-aef6-4ed9-bdc9-705f59ab9ad6} & Rotate a video and save it. \\
\texttt{5ac2891a-eacd-4954-b339-98abba077adb} & Disable automatic exit after playback. \\
\addlinespace
\multicolumn{2}{@{}l}{\textbf{VS Code (3)}}\\*
\texttt{eabc805a-bfcf-4460-b250-ac92135819f6} & Install the Python extension. \\
\texttt{930fdb3b-11a8-46fe-9bac-577332e2640e} & Customize keyboard shortcuts (advanced configuration). \\
\texttt{7c4cc09e-7a92-40dd-8338-b2286535c4ed} & $\dagger$ Switch the interface to Arabic without extensions. \\
\addlinespace
\end{longtable}
\endgroup

\clearpage
\subsection{Composition of the Controlled Subset}
\label{app:subset}
Table~\ref{tab:subset_coverage} reports the fixed, non-random subset used in Appendices~\ref{sec:controlled_studies} and~\ref{sec:inference_cost}.
The subset is selected from the same 361-task evaluation set used for the main results and the same-backbone Sonnet~4.6 comparison. Coverage is calculated against this set, which excludes the eight Google Drive tasks.
The ten groups below are application-level groups, rather than the five aggregated reporting categories in Table~\ref{tab:main_results}.

\begin{table}[htbp]
\centering
\small
\caption{Application-domain coverage of the fixed 60-task subset within our 361-task OSWorld evaluation set.}
\label{tab:subset_coverage}
\begin{tabular}{@{}lrrr@{}}
\toprule
Domain group & Subset tasks & Evaluation-set tasks & Coverage (\%) \\
\midrule
Chrome & 8 & 46 & 17.4 \\
GIMP & 4 & 26 & 15.4 \\
LibreOffice Calc & 8 & 47 & 17.0 \\
LibreOffice Impress & 8 & 47 & 17.0 \\
LibreOffice Writer & 4 & 23 & 17.4 \\
Multi-app & 15 & 93 & 16.1 \\
OS & 4 & 24 & 16.7 \\
Thunderbird & 3 & 15 & 20.0 \\
VLC & 3 & 17 & 17.6 \\
VS Code & 3 & 23 & 13.0 \\
\midrule
Total & 60 & 361 & 16.6 \\
\bottomrule
\end{tabular}
\end{table}

\subsection{Inference Cost and Tool Attribution}
\label{sec:inference_cost}

Table~\ref{tab:usage_subset} reports mean environment actions and tracked usage per task on the same 60-task subset, using only the 100-step
 Manager-budget configuration. The action counts are observed means, not separate evaluation budgets.
The model-call accounting includes executor, Reflection, Coder, Gate, and Judge records where tracked; these are usage-log categories and do
 not imply that every Gate check is an independent model call.
The $\geq$ markers are retained because token/cost coverage, and call coverage for some variants, is incomplete. Tracked cost is therefore a lower bound on API expenditure rather than a complete end-to-end cost estimate.
Separate input/output-token totals, complete provider/tool charges, and infrastructure costs are not available in these measurements.

\begin{table}[tbp]
\centering
\footnotesize
\setlength{\tabcolsep}{4pt}
\caption{Mean usage per task on the fixed 60-task subset with a 100-step Manager budget. All scores use this budget; calls and observed
 actions are different quantities. Dollar values are tracked API cost in USD; $\geq$ denotes a lower bound from incomplete usage accounting.}
\label{tab:usage_subset}
\begin{tabular}{@{}lrrrrr@{}}
\toprule
Configuration & Score (\%) & Actions & Model calls & Tokens & Cost (\$) \\
\midrule
Base scaffold & 65.05 & 16.80 & 28.40 & $\geq257{,}618$ & $\geq0.137$ \\
+ Search Agent & 68.86 & 12.70 & $\geq32.55$ & $\geq343{,}640$ & $\geq0.227$ \\
+ Completeness-Aware Planner & 67.19 & 27.32 & $\geq89.13$ & $\geq901{,}936$ & $\geq0.717$ \\
+ Loop Breaker & 69.00 & 14.28 & $\geq41.52$ & $\geq422{,}074$ & $\geq0.323$ \\
Full & \textbf{77.15} & 19.67 & 47.39 & $\geq579{,}719$ & $\geq0.316$ \\
\bottomrule
\end{tabular}
\end{table}

Among the standalone additions, Completeness-Aware Planner has the largest tracked overhead and increases mean environment actions from
 16.80 to 27.32.
The full configuration achieves a higher score with fewer tracked calls than the CAP-only configuration, indicating that overhead is not
 additive across modules because the modules change the execution trajectory.
The reported cost lower bounds do not support a ranking of complete API expenditure.
Accordingly, performance under a fixed Manager-action budget should not be interpreted as equivalent model-call, latency, or monetary
 efficiency.

\subsection{Broader Impact and Limitations}
\label{app:limitations}

\paragraph{Models, tools, and inference overhead}
All evaluated Manager backbones are proprietary. The additional Gemini~3~Flash studies also use proprietary Gemini~3.1~Pro Coder and Searcher models when present, so they do not establish generality to open models.
Model-version changes, API availability, and changing external search results limit exact reproducibility.
Coding and grounding are adopted tools, and their invocation counts do not replace independent access-matched ablations.

\paragraph{Evaluation and remaining verification failures}
The fixed 60-task subset is non-random; it supplies controlled point estimates, not population-level or statistically significant claims.
Loop-control thresholds are hand-designed and have not been evaluated in a sensitivity sweep.
The same-backbone and transfer experiments reduce confounding by Manager capability, but cover only the reported models and scaffolds.
False completion remains dominant among failures, and visually plausible states can still conceal incorrect file contents, pending writes, or evaluator-relevant state.

\paragraph{Deployment safeguards}
Our results come from benchmark virtual machines, not unrestricted user desktops.
Incorrect or overconfident actions can modify, delete, send, or expose information before a stopping error is detected.
Practical deployment requires sandboxing, scoped permissions, confirmation before irreversible or sensitive operations, audit logs, command allowlists, and rollback mechanisms.
Completion verification is a reliability mechanism and does not itself provide these security or authorization controls.

\paragraph{Limitations and Future directions}
Our current system adopts a relatively simple memory and planning architecture without long-horizon task decomposition or cross-task
 knowledge transfer. More sophisticated memory mechanisms may further improve performance on complex multi-step tasks.
  On the other hand, \name generates high-quality, verified execution trajectories as a byproduct of its verification and loop recovery.
   A promising direction is leveraging these trajectories to train unified end-to-end multimodal models, distilling agentic reasoning into
    a single model that acts directly from pixels---bridging the reliability of agentic frameworks with the efficiency of end-to-end
     approaches.

\subsection{Implementation Details}
\label{app:impl}

\subsubsection{Configurations.}
Table~\ref{tab:model_config} summarizes the model assignments used for each component across our experimental configurations.
For the main OSWorld evaluation, we use six configurations. Three use Claude Opus~4.5, Claude Sonnet~4.6, or Gemini~3~Flash as the shared backbone for the Manager Agent, Reflection Agent, Completeness Model Judge, and Coding Agent, with Gemini~3~Pro~\citep{gemini3pro2025} as the Search Agent. We additionally evaluate an Opus~4.5 variant that keeps the same backbone and Search Agent but replaces the grounding model with MAI-UI, an Opus~4.6~\citep{claudeopus46} configuration that uses Gemini~3.1~Pro~\citep{gemini31pro2025} as the Search Agent, and a Gemini~3.1~Pro configuration that uses Gemini~3.1~Pro across all components except grounding. On WindowsAgentArena, we evaluate Gemini~3~Flash as the Manager backbone.

\begin{table}[h]
\centering
\scriptsize
\caption{Model assignments for the main-suite configurations. The additional 60-task studies instead use Gemini~3.1~Pro for the Coder and Searcher when present.}
\label{tab:model_config}
\begin{tabularx}{\textwidth}{
    >{\raggedright\arraybackslash}p{0.28\textwidth}
    >{\raggedright\arraybackslash}X
    >{\raggedright\arraybackslash}p{0.18\textwidth}
}
\toprule
\textbf{Component} & \textbf{Model} & \textbf{Provider} \\
\midrule
Manager Agent & Claude Opus~4.5 / Claude Opus~4.6 / Sonnet~4.6 / Gemini~3~Flash / Gemini~3.1~Pro  & Anthropic / Google \\
Reflection Agent (Loop Breaker) & Same as the Manager Agent & Anthropic / Google \\
Completeness Model Judge & Same as Manager Agent & Anthropic / Google \\
Search Agent & Gemini~3~Pro for Opus~4.5 / Opus~4.5 + MAI-UI / Sonnet~4.6 / Gemini~3~Flash; Gemini~3.1~Pro for Opus~4.6 / Gemini~3.1~Pro (w/ search grounding) & Google \\
Coding Agent & Same as the Manager Agent & Anthropic / Google \\
Grounding Agent & Seed~1.8 for all configurations except MAI-UI for Opus~4.5 + MAI-UI & ByteDance / -- \\
\bottomrule
\end{tabularx}
\end{table}

\subsubsection{Hyperparameters.}
All agents operate in iterative planning mode (no hierarchical subtask decomposition). The per-task step budget
 is set to 15, 50, or 100 actions depending on the experiment. The Completeness Model Judge within
  Completeness-Aware Planner uses temperature $T{=}0.2$ for conservative judgment; all other
   components use $T{=}1.0$. The Coding Agent has an independent step budget of 20 code execution steps.
    The Search Agent performs a single query.

\subsubsection{Action-Outcome Verification Rules.}
Table~\ref{tab:micro_verify} lists the rule-based checks Completeness-Aware Planner applies to the expected visual outcome of each action type.

\begin{table}[h]
\centering
\small
\caption{Verification rules by action type in Completeness-Aware Planner.}
\label{tab:micro_verify}
\begin{tabular}{@{}ll@{}}
\toprule
\textbf{Action Type} & \textbf{Expected Verification} \\
\midrule
Click button/menu & New UI element visible (dialog, tab, highlight) \\
Toggle setting & State label changed (``Enable'' $\to$ ``Disable'') \\
Type text & Field contains typed text; cursor moved \\
Export/save & New file in folder, success toast, or title bar change \\
No visible change & \texttt{wait(1)} before re-check; do not repeat immediately \\
\bottomrule
\end{tabular}
\end{table}

\subsubsection{Infrastructure.}
For OSWorld, experiments run on the officially released Ubuntu Docker containers hosted on Amazon Web Services. 
For WindowsAgentArena, we use the official Windows environments infrastructure and deploy the environment snapshot on Amazon Web Services virtual machines. 
Screenshots are captured at 1920$\times$1080 resolution and served as the primary observation modality.

\subsubsection{Reported Categories.}
On OSWorld, we follow the official evaluation setting, where tasks are grouped into the following categories: OS, Office (LibreOffice Calc, Impress, Writer), Daily (Chrome, VLC Player, Thunderbird), Professional (VS Code and GIMP), and Workflow (tasks involving interactions across multiple applications).
On WindowsAgentArena, tasks are organized into several functional categories following OS-Symphony. Office includes LibreOffice Writer and LibreOffice Calc tasks; Web (Web Browsing) includes Edge and Chrome tasks; Sys. (Windows System) includes File Explorer and Settings tasks; Code includes VSCode tasks; Media (Media \& Video) includes VLC tasks; and Util. (Windows Utilities) includes Notepad, Clock, Paint, and WindowsCalc tasks.

\subsection{Action Space}
\label{app:action_space}

Table~\ref{tab:action_space} lists the executable actions exposed to the Manager Agent in the full-system configuration. The table shows the Linux/OSWorld setting with the Coding Agent and Search Agent enabled. The \texttt{save\_to\_knowledge} action (for memory) is excluded in this configuration.

\begin{table}[t]
\centering
\scriptsize
\caption{Action space exposed to the Manager Agent in the full-system configuration.}
\label{tab:action_space}
\begin{tabularx}{\textwidth}{
    >{\raggedright\arraybackslash}p{0.32\textwidth}
    >{\raggedright\arraybackslash}p{0.14\textwidth}
    X
}
\toprule
\textbf{Action} & \textbf{Type} & \textbf{Description} \\
\midrule
\nolinkurl{agent.click} & GUI & Clicks a grounded UI element with optional repeated clicks, mouse-button choice, and modifier keys. \\
\nolinkurl{agent.double_click} & GUI & Double-clicks a grounded UI element. \\
\nolinkurl{agent.type} & GUI & Focuses an element if needed, types text, optionally overwrites existing text, and can submit with Enter. \\
\nolinkurl{agent.drag_and_drop} & GUI & Drags from one grounded element to another grounded element. \\
\nolinkurl{agent.highlight_text_span} & GUI & Selects text between a starting phrase and an ending phrase. \\
\nolinkurl{agent.scroll} & GUI & Scrolls vertically or horizontally over a grounded target region. \\
\nolinkurl{agent.open} & Navigation & Opens an application or file. \\
\nolinkurl{agent.switch_applications} & Navigation & Switches to an already opened application. \\
\nolinkurl{agent.hotkey} & Keyboard & Sends a key chord such as \texttt{Ctrl+S}. \\
\nolinkurl{agent.hold_and_press} & Keyboard & Holds one set of keys while pressing another key sequence. \\
\nolinkurl{agent.call_code_agent} & Tool & Delegates a task or subtask to the Coding Agent, which runs Python/Bash and returns a structured report. \\
\nolinkurl{agent.call_search_agent} & Tool & Calls the Search Agent to retrieve a step-by-step tutorial for a specific GUI procedure. \\
\nolinkurl{agent.set_cell_values} & Tool & Performs spreadsheet cell edits programmatically (Linux-specific). \\
\nolinkurl{agent.wait} & Terminal & Waits for the UI to update before the next decision. \\
\nolinkurl{agent.done} & Terminal & Terminates the task successfully. \\
\nolinkurl{agent.fail} & Terminal & Terminates the task as infeasible or blocked. \\
\bottomrule
\end{tabularx}
\end{table}

\subsection{System Prompts for Agents}
\label{app:prompts}
\subsubsection{Reflection Agent Prompt (Loop Breaker Tier~3)}
\label{app:reflection}

The Reflection Agent serves as the external model judge in the Loop Breaker's Tier~3 (Section~\ref{sec:loop_breaker}). It is invoked after every Manager action to analyze the trajectory and produce structured signals. When the Strategy signal is \textsc{switch}, the Loop Breaker injects a hard directive forcing the Manager to change its approach. The full prompt is shown below.

\begin{Verbatim}[
    frame=single,
    framesep=2mm,
    rulecolor=\color{black},
    fontsize=\scriptsize,
    breaklines=true
]
You are a reflection agent designed to assist in task execution by
reflecting on the trajectory of a task and providing feedback for
what the next step should be.
You have access to the Task Description and the Current Trajectory
of another computer agent. The Current Trajectory is a sequence of a
desktop image, chain-of-thought reasoning, and a desktop action for
each time step. The last image is the screen's display after the last
action.
Your task is to generate a reflection. Your generated reflection must
fall under one of the two cases listed below:

Case 1. The trajectory is not going according to plan. This is often
due to the latest action not being executed correctly, or a cycle of
actions being continually repeated with no progress being made. In this
case, explicitly highlight why the current trajectory is incorrect, and
encourage the computer agent to try a new action. However, DO NOT
encourage a specific action in particular.

Case 2. The trajectory is going according to plan. In this case, simply
tell the agent to continue proceeding as planned. DO NOT encourage a
specific action in particular.

To be successful, you must follow the rules below:
- DO NOT suggest any specific future plans or actions. Your only goal
  is to provide a reflection, not an actual plan or action.
- Your reflection MUST be evidence-based and outcome-focused, not
  procedural.
- Any response that falls under Case 1 should explain why the
  trajectory is not going according to plan. You should especially look
  out for cycles of actions that are continually repeated with no
  progress.
- Any response that falls under Case 2 should be concise, since you
  just need to affirm the agent to continue with the current
  trajectory.

Required fields (write them as labeled lines):
- Progress signal: What visibly changed since the prior step?
  (or "no visible change")
- Outcome signal: Did we achieve the intended subgoal (not just click
  something)? What is the evidence?
- Loop signal: Are we repeating the same interaction/attempt 2+ times
  or seeing the same screen 3 times? (yes/no + evidence)
- Feasibility signal: Is the task still feasible?
  (feasible / uncertain / impossible + evidence)
- Termination signal: DONE / FAIL / CONTINUE (evidence-based). Use
  DONE only when ALL task success criteria have SPECIFIC VISIBLE
  EVIDENCE. Use FAIL when the task is demonstrably infeasible. Use
  CONTINUE otherwise.
- Strategy signal: KEEP / SWITCH (and brief reason if SWITCH). If
  Loop signal = yes OR Feasibility signal = uncertain/impossible,
  Strategy signal should usually be SWITCH.
- Verdict: Case 1 or Case 2
\end{Verbatim}

\subsubsection{Completeness-Aware Planner: Model Judge Prompt}
\label{app:verifier}

The Completeness Model Judge within Completeness-Aware Planner is invoked whenever the Manager's Completion Gate outputs \textsc{done} (Section~\ref{sec:completion_aware_planning}). It receives the task instruction, current screenshot, and recent trajectory, and produces a binary accept/reject decision. The prompt enforces conservative verification: it requires direct visual evidence for every criterion and prefers false negatives over false positives. The full prompt is shown below.

\begin{Verbatim}[
    frame=single,
    framesep=2mm,
    rulecolor=\color{black},
    fontsize=\scriptsize,
    breaklines=true
]
You are an extremely strict task completion verifier for desktop
automation. Your job is to decide whether the task is ALREADY fully
completed right now.

Decision rules (be conservative):
- Mark completion ONLY if every requirement in the instruction is
  satisfied AND there is direct, unambiguous evidence in the current
  screenshot and/or provided recent history.
- Do NOT assume hidden state or intent. If something could be
  incomplete, mark incomplete.
- For side effects (sending/submitting/saving/downloading/creating/
  deleting/installing/moving), require visible confirmation (success
  toast, item present in the right place, sent email visible, saved
  file visible, etc.).
- Require a stable UI state when possible (no open dropdowns/menus,
  no blocking modal, no loading indicator suggesting the action is
  still in progress).
- Prefer false negatives over false positives.

Semantic evidence rules (prevent false positives from workarounds):
- If the task involves a file format requirement (e.g., "export as
  SVG", "save as PDF"), require visible evidence of the correct file
  extension AND an app dialog/status/file-browser entry confirming
  that format. If not visible, mark incomplete.
- If the task involves a mode or state (e.g., "CMYK mode", "dark
  theme enabled"), require explicit UI indicators (mode label in
  title/status bar, toggle state). Do NOT infer from visual appearance
  alone.
- If a workaround was used instead of the exact requested method,
  verify it produces a semantically equivalent result. If equivalence
  is uncertain, mark incomplete.

Exact-value verification (CRITICAL):
- If the task specifies an EXACT value (hex color, font size, config
  key, URL parameter, file path, specific text string, numeric value),
  you MUST find that EXACT value readable on screen or confirmed in
  the trajectory text.
- "Looks like the right color" does NOT verify "#00FF00". "Looks like
  a small font" does NOT verify "11pt".
- For values set via code agent: check the trajectory for the code
  agent's output confirming the value was written/set correctly.
- For values set via GUI: the value must be readable in an input
  field, status bar, dialog, or properties panel on the current
  screenshot.
- If the exact value cannot be confirmed, mark INCOMPLETE.

Trajectory cross-check (MANDATORY):
- Review the trajectory text for actual actions taken. Verify that the
  actions match what the task requested.
- If the code agent ran commands, check the command output in the
  trajectory for success/failure indicators and correct values.
- If the trajectory shows errors, failed commands, or values different
  from what was requested, mark INCOMPLETE regardless of how the
  screenshot looks.

Output:
- Return ONLY a JSON object with:
  {"complete": true/false, "reason": "<short>",
   "missing_steps": "<what is still needed, or empty if complete>"}
\end{Verbatim}

The verifier agent also applies conservative post-processing: if the model claims completion but lists missing steps or uses uncertainty phrases (\eg, ``not sure'', ``unclear'', ``cannot verify''), the verdict is automatically overridden to incomplete.

\subsubsection{Manager Agent System Prompt}
\label{app:system_prompt}

Below we report the full prompt for the Manager agent under the default Linux configuration: (1)~a termination contract that mandates
 explicit terminal actions, (2)~the Completion Gate protocol enforcing per-step self-verification, (3)~micro-verification rules for
  post-action checks, (4)~the Loop Breaker rules (Tiers~1--2), (5)~feasibility management guidelines, (6)~agent usage guidelines
   describing when to invoke the Coding and Search Agents, and (7)~the output format specification.

\begin{Verbatim}[
    frame=single,
    framesep=2mm,
    rulecolor=\color{black},
    fontsize=\tiny,
    breaklines=true
]
You are an expert in graphical user interfaces and Python code. You are
responsible for executing the task: `TASK_DESCRIPTION`.
You are working in linux.

# GUIDELINES

## Termination Contract (NON-NEGOTIABLE)
Every trajectory MUST terminate with exactly one terminal API call:
- `agent.done()` when ALL success criteria are visibly satisfied.
- `agent.fail()` when the task is infeasible or permanently blocked.
NEVER "run out" the episode without a terminal call. If uncertain,
perform a verification action, then decide.
If success criteria are satisfied, the very next grounded action MUST
be `agent.done()` -- no extra cleanup steps.

## Completion Gate (run FIRST every step)
Before choosing any action, make a binary decision using only visible
evidence from the current observation and recent history:
1. (Step 1 only, or when the task changes): Write 1-3 success criteria
   as UI-observable statements (button labels, file names in lists,
   toggle states, presence/absence of UI elements, dialog titles). If
   a criterion is not directly observable, rewrite it into an
   observable proxy. If a criterion requires an exact value (hex color,
   font size, config key, file content), tag it as [EXACT CHECK].
2. (Every step): For each criterion, update its status using ONLY
   visible evidence. For [EXACT CHECK] criteria: cite the literal
   value readable on screen or confirmed by code agent output.
3. If ALL criteria are satisfied AND the UI is stable, the next
   grounded action MUST be `agent.done()`.
4. If the task is impossible based on evidence, the next grounded
   action MUST be `agent.fail()`.
5. Otherwise, proceed with one concrete next action.

## Micro-Verification (MANDATORY after every UI action)
When verifying the previous action, apply the appropriate check:
- Clicked a button/menu: Verify the expected new UI element is visible.
- Toggled a setting: Verify the toggle state actually changed.
- Typed text: Verify the field contains the typed text.
- Exported/saved: Verify the new file appears or success toast shows.
- No visible change: Do NOT repeat. Wait, then re-check or switch
  strategy.

## Hard Loop Breaker (MANDATORY)
- If the same action type + same target produces no visible change
  twice, the next action MUST be a different modality.
- If the same screen state appears three times in a row with no
  progress, you MUST switch strategy completely. Call `agent.fail()`
  only when separate feasibility evidence establishes infeasibility.
- When the reflection agent returns Strategy signal = SWITCH, you MUST
  change modality/approach on the very next action.

## Feasibility Management
If the instruction depends on uncertain prerequisites:
1. Identify the 1-3 most likely prerequisites.
2. Run a short feasibility probe (first 1-3 steps).
3. Maintain a small evidence trail of what you checked.
4. If after ~6-8 actions there is no positive evidence, call
   `agent.fail()`.

## Agent Usage Guidelines
You have access to: GUI, Code, Search. Choose the correct one.

### GUI Agent
- Use for: All direct UI interactions (clicking, typing, dragging).

### Code Agent
Use code agent ONLY when:
- The task requires bulk edits (>= 20 cells/lines), OR
- The task requires non-trivial computation, OR
- The GUI route is blocked.
Do NOT use code agent when the change is achievable in <= 3 GUI
actions or for charts/graphs/pivot tables/visual layout.

### Search Agent
- Use when you are unsure how to perform a GUI-based task.
- Call with a clear "How to" query for a single specific action.

# Output Format
You are provided with: (1) A screenshot, (2) History of previous
interactions, (3) A text reflection from the Reflection Agent,
(4) Tutorials from the Search Agent.

Your response must include:
(Completion Gate) - success criteria, evidence check, feasibility
status, and decision (DONE / FAIL / CONTINUE).
(Previous action verification) - micro-verification of the last action.
(Screenshot Analysis) - description of current desktop state.
(Next Action) - natural language description of the next action.
(Grounded Action) - a single Python API call, e.g.:
```python
agent.click("The menu button at the top right", 1, "left")
```
\end{Verbatim}

\subsection{Additional Ablation Results}
\label{app:ablations}

This section provides detailed per-budget breakdowns for Completeness-Aware Planner and Loop Breaker, supplementing the aggregate analysis in the main paper.

\subsubsection{Stronger Backbones Leverage Tools More Efficiently.}
\begin{figure}[t]
    \centering
    \includegraphics[width=1\linewidth]{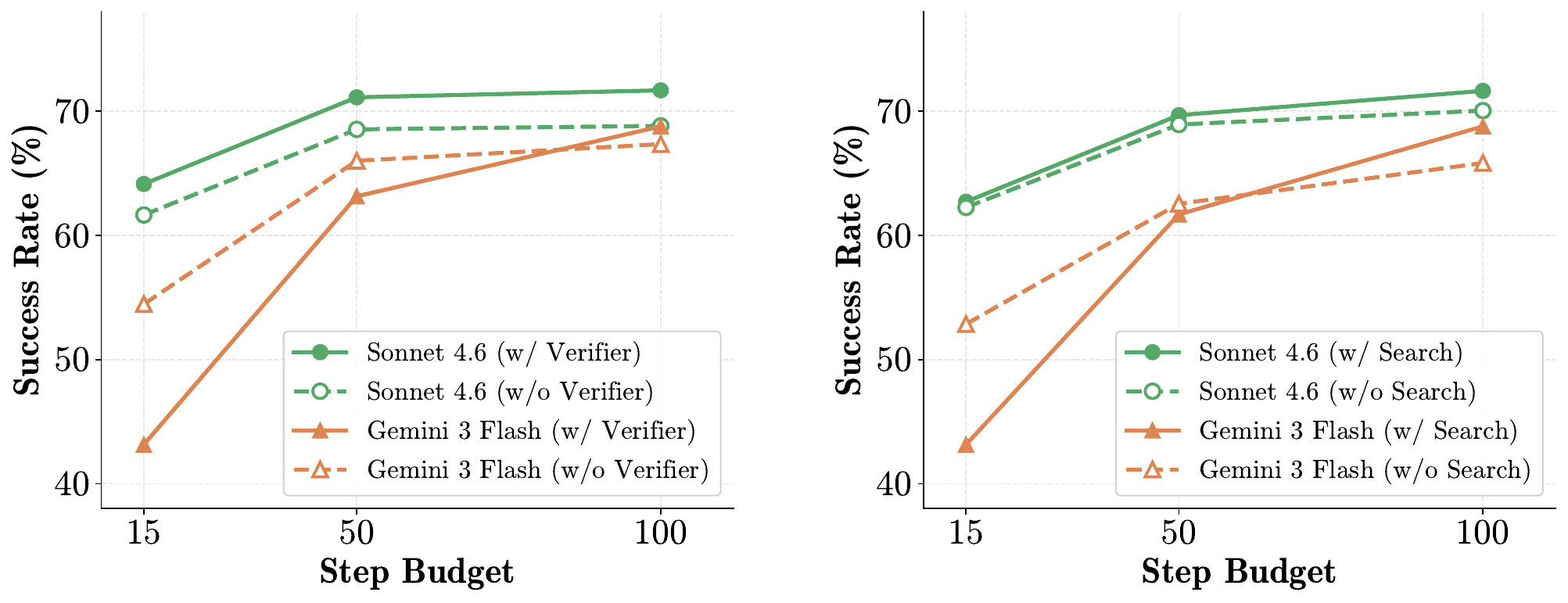}
     \caption{Impact of Completeness-Aware Planner and Search Agent across step budgets on OSWorld. Sonnet~4.6 benefits consistently at all budgets, while Gemini~3~Flash gains only at relaxed budgets: tool calls consume actions that less efficient models cannot afford under tight constraints.}
     \label{fig:component_ablation1}
   \end{figure}

Figure~\ref{fig:component_ablation1} examines how Completeness-Aware Planner and Search Agent interact with the step budget across both Claude Sonnet 4.6 and Gemini 3 Flash.
For Sonnet~4.6, both components deliver consistent gains at every budget (\eg, CAP +2.5\% at 15 steps, +2.9\% at 100 steps), as Sonnet is step-efficient enough to absorb the action overhead of tool invocations.
When switching to Gemini~3~Flash, both components help at 100 steps (\eg, Search +3.0\%) but hurt under tighter budgets (\eg, CAP $-$11.3\% and Search $-$9.7\% at 15 steps). 
This occurs because invoking tools inherently consumes action steps. Gemini 3 Flash already needs more raw actions to complete tasks, so the tool overhead crowds out task execution under constrained budgets.

\subsubsection{Completeness-Aware Planner: Failure Breakdown across Step Budgets}
\label{app:verifier_ablation}

Table~\ref{tab:verifier_breakdown} reports the false completion count and DONE accuracy across step budgets for both backbones, with and without Completeness-Aware Planner. \textit{False DONE} is the number (and percentage over all failed cases) of tasks where the agent incorrectly declared success. \textit{DONE Accuracy} is the percentage of all DONE signals that are correct.

\begin{table}[h]
\centering
\scriptsize
\caption{False completion breakdown across step budgets on OSWorld. \textbf{w/}: full system with Completeness-Aware Planner; \textbf{w/o}: CAP removed. False DONE reports the count and fraction of failed tasks with false completion. DONE Accuracy reports the fraction of all completion signals that are correct.}
\label{tab:verifier_breakdown}
\begin{tabular}{@{}l l cc cc cc@{}}
\toprule
& & \multicolumn{2}{c}{\textbf{@100 steps}} & \multicolumn{2}{c}{\textbf{@50 steps}} & \multicolumn{2}{c}{\textbf{@15 steps}} \\
\cmidrule(lr){3-4} \cmidrule(lr){5-6} \cmidrule(lr){7-8}
\textbf{Backbone} & \textbf{Metric} & w/ & w/o & w/ & w/o & w/ & w/o \\
\midrule
\multirow{2}{*}{Sonnet~4.6}
& False DONE & 91 (91.9\%) & 105 (95.5\%) & 86 (85.1\%) & 98 (88.3\%) & 63 (50.0\%) & 73 (53.7\%) \\
& DONE Acc. & 73.5\% & 69.6\% & 74.4\% & 70.9\% & 78.2\% & 74.7\% \\
\midrule
\multirow{2}{*}{Gemini~3~Flash}
& False DONE & 100 (86.2\%) & 102 (91.9\%) & 72 (52.6\%) & 97 (80.2\%) & 28 (13.7\%) & 63 (38.7\%) \\
& DONE Acc. & 70.3\% & 69.8\% & 74.6\% & 70.6\% & 84.3\% & 75.5\% \\
\bottomrule
\end{tabular}
\end{table}

For Sonnet~4.6, Completeness-Aware Planner consistently reduces the false DONE rate across all budgets, with DONE accuracy improving by 3--4\% at every tier (\eg, 69.6\%$\to$73.5\% at 100 steps). The effect is moderate because Sonnet already produces relatively calibrated completion signals.

For Gemini~3~Flash, the verifier has a much larger effect under tighter budgets. At 50 steps, the false DONE fraction drops sharply from 80.2\% to 52.6\% ($-$27.6\%), and DONE accuracy rises from 70.6\% to 74.6\%. At 15 steps the gap is even starker: 38.7\%$\to$13.7\% ($-$25.0\%). This is because without the verifier, Flash tends to declare completion early when it cannot make progress within the tight budget; the verifier catches these premature claims. At 100 steps, where Flash has enough budget to actually complete more tasks, the gap narrows (91.9\%$\to$86.2\%).

\subsubsection{Loop Breaker: Effectiveness across Step Budgets}
\label{app:loop_ablation}

Table~\ref{tab:loop_budget} reports task success rates with and without the Loop Breaker across step budgets for both backbones.

\begin{table}[h]
\centering
\small
\caption{Loop Breaker ablation across step budgets on OSWorld. $\Delta$ is the difference when adding the Loop Breaker.}
\label{tab:loop_budget}
\begin{tabular}{@{}l ccc@{}}
\toprule
\textbf{Budget} & \textbf{w/ Loop Breaker} & \textbf{w/o Loop Breaker} & \textbf{$\Delta$} \\
\midrule
\multicolumn{4}{l}{\textit{Sonnet~4.6}} \\
\quad @100 & 71.67\% & 71.63\% & +0.04\% \\
\quad @50  & 71.11\% & 69.67\% & +1.44\% \\
\quad @15  & 64.13\% & 62.70\% & +1.43\% \\
\midrule
\multicolumn{4}{l}{\textit{Gemini~3~Flash}} \\
\quad @100 & 68.77\% & 66.95\% & +1.82\% \\
\quad @50  & 63.14\% & 58.90\% & +4.24\% \\
\quad @15  & 43.15\% & 49.30\% & $-$6.15\% \\
\bottomrule
\end{tabular}
\end{table}

For Sonnet~4.6, the Loop Breaker provides consistent gains at tighter budgets (+1.4\% at both 15 and 50 steps), while the effect at 100 steps is negligible (+0.04\%). This aligns with the main paper's finding that Sonnet loops infrequently---at 100 steps it has enough budget to self-recover, so explicit loop breaking adds little; under tighter budgets, even occasional loops waste precious steps that the Loop Breaker reclaims.
For Gemini~3~Flash, the Loop Breaker helps substantially at 100 steps (+1.82\%) and 50 steps (+4.24\%), where it loops frequently and the escalation mechanism redirects the agent toward productive actions. However, at 15 steps the Loop Breaker \emph{hurts} ($-$6.15\%). The Loop Breaker's escalation tiers (modality switch, strategy change) themselves consume actions, and at 15 steps Gemini~3~Flash cannot afford this overhead. When budget is sufficient, the cost of escalation is recouped through recovered trajectories; when budget is tight, the escalation actions crowd out task execution.

\end{document}